\documentclass[10pt,twocolumn,letterpaper]{article}

\usepackage{cvpr}              
\usepackage{caption}
\usepackage[dvipsnames]{xcolor}

\usepackage[pagebackref,breaklinks,colorlinks,citecolor=cvprblue]{hyperref}
\usepackage[capitalize]{cleveref}
\crefname{section}{Sec.}{Secs.}
\Crefname{section}{Section}{Sections}
\Crefname{table}{Table}{Tables}
\crefname{table}{Tab.}{Tabs.}

\renewcommand{\paragraph}[1]{\medskip\noindent\textbf{#1}}
\newcommand{\xmark}{\ding{55}}%

\usepackage{times}
\usepackage{epsfig}
\usepackage{graphicx}
\usepackage{amsmath}
\usepackage{amssymb}
\usepackage{graphicx}
\usepackage{amsmath}
\usepackage{amssymb}
\usepackage{booktabs}
\usepackage{pifont}
\usepackage{soul}
\usepackage{xcolor}
\usepackage{caption}
\usepackage{xspace}

\usepackage{algorithm,algpseudocode}
\algnewcommand{\LeftComment}[1]{\Statex \(\triangleright\) #1}
\newcommand{\startemplate}{\ensuremath{T_\mathit{SMPL}}\xspace}
\newcommand{\pcashapespace}{\ensuremath{\mathcal{B}_\mathit{PCA}}\xspace}
\newcommand{\regipca}{\ensuremath{\mathcal{R}_\mathit{PCA}}\xspace}
\newcommand{\reginjf}{\ensuremath{\mathcal{R}_\mathit{DEFORM}}\xspace}

\newcommand{\losstotal}{\ensuremath{L_\mathit{total}}\xspace}
\newcommand{\lossvert}{\ensuremath{L_\mathit{vertex}}\xspace}
\newcommand{\lossjacob}{\ensuremath{L_\mathit{Jacobian}}\xspace}

\newcommand{\scans}{\ensuremath{\mathcal{S}}\xspace}
\newcommand{\regiset}{\ensuremath{\mathcal{R}}\xspace}
\newcommand{\unregiset}{\ensuremath{\mathcal{U}}\xspace}
\newcommand{\regi}{\ensuremath{X}\xspace}
\newcommand{\scan}{\ensuremath{S}\xspace}
\newcommand{\scanofregiset}{\ensuremath{\mathcal{S}_R}\xspace}
\newcommand{\scanofregi}{\ensuremath{S_X}\xspace}
\newcommand{\pcaproj}{\ensuremath{X_o}\xspace}
\newcommand{\njfresult}{\ensuremath{X'}\xspace}
\newcommand{\canonical}{\ensuremath{S_c}\xspace}
\newcommand{\posed}{\ensuremath{S_p}\xspace}
\newcommand{\pose}{\ensuremath{\theta}\xspace}

\newcommand{\totalartistshape}{429\xspace}

\newcommand{\oursvtwo}{\name\xspace}

\newcommand{\baselineB}{\textsc{FullAnnotation}\xspace}
\newcommand{\baselineC}{\textsc{Baseline1}\xspace}
\newcommand{\baselineD}{\textsc{Baseline2}\xspace}

\newcommand{\oursvtwopcaonly}{\name (PCA only)\xspace}

\newcommand{\numunregiscanVtwo}{800\xspace}

\newcommand{\regieval}{\ensuremath{\mathcal{R}_\mathit{EVAL}}\xspace}

\newcommand{\baselinefullpcanjf}{\baselineB - PCA+NJF\xspace}

\definecolor{peach}{RGB}{255,145,145}

\newcommand{\name}{BLiSS\xspace}

\definecolor{cvprblue}{rgb}{0.21,0.49,0.74}

\def\paperID{0179} 
\def\confName{3DV\xspace}
\def\confYear{2024\xspace}

\begin{document}
\title{\vspace{-1 em}\name: Bootstrapped Linear Shape Space\vspace{-1 em}}

\author{
  \href{https://sanjeevmk.github.io/}{Sanjeev Muralikrishnan}\textsuperscript{1}\\
  \and
  \href{https://research.adobe.com/person/paulchhuang/}{Chun-Hao Paul Huang}\textsuperscript{2} \\
  \and
  \href{https://www.duygu-ceylan.com/}{Duygu Ceylan}\textsuperscript{2} \\
    \and
  \href{http://www0.cs.ucl.ac.uk/staff/n.mitra/}{Niloy J. Mitra}\textsuperscript{1,2} \\
}

\twocolumn[{%
\renewcommand\twocolumn[1][]{#1}%
\maketitle
\vspace{-.4in}
\begin{center}
\begin{tabular}{c@{\hspace{0.5cm}}c}
    \large\textsuperscript{1} University College London & \large\textsuperscript{2} Adobe Research
\end{tabular}
\end{center}
\begin{center}
  \large {\href{https://geometry.cs.ucl.ac.uk/group_website/projects/2023/bliss/}{Project Webpage}}
\end{center}

\begin{center}
\centering
     \includegraphics[width=1\linewidth]{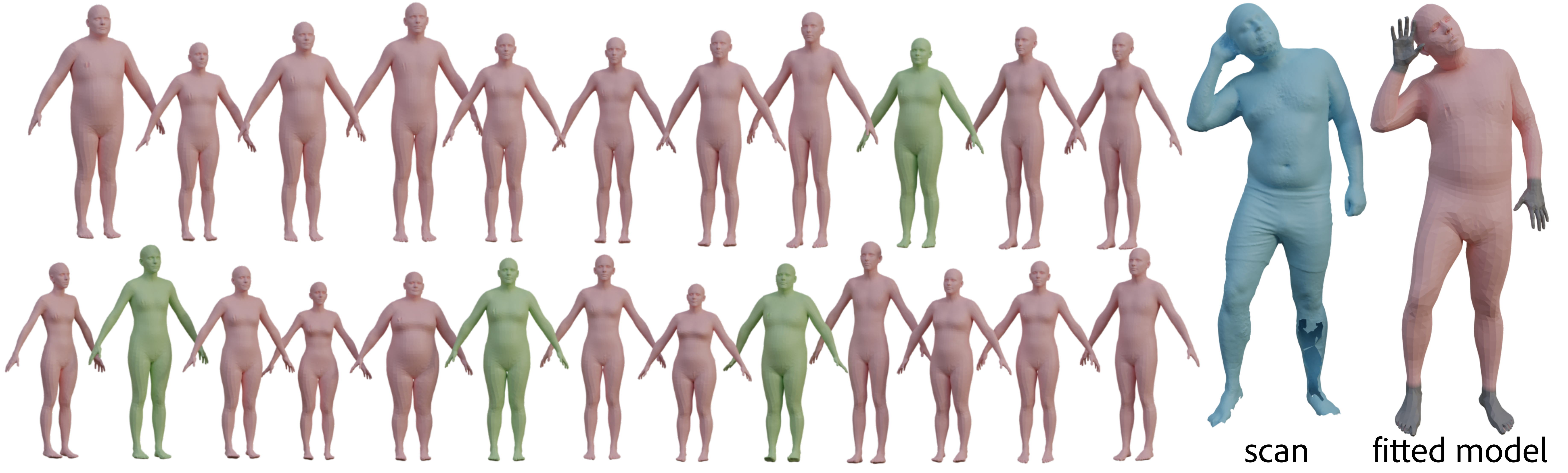}
     \vspace{-.2in}
     \captionsetup{type=figure}
     \caption{We present \name, which progressively builds a  human body shape space and brings {\em unregistered} scans into correspondence to a given template mesh. Starting from as few as $200$ manually registered scans (green samples), \name creates an expressive shape space (pink samples), performing on par with state-of-the-art models such as SMPL, STAR, and GHUM, while requiring only 5\% of annotations compared to the others. 
(Right)~Our space can then recover the body-shape parameters of raw scans by projecting them directly to ours.}
    \label{fig:teaser}
    \vspace{+0.50em}
\end{center}
}]

\begin{abstract}

Morphable models are fundamental to numerous human-centered processes as they offer a simple yet expressive shape space. 
Creating such morphable models, however, is both tedious and expensive. 
%
%
The main challenge is establishing dense correspondences across raw scans that capture sufficient shape variation. This is often addressed using a mix of significant manual intervention and non-rigid registration.
We observe that creating a shape space and solving for dense correspondence are tightly coupled -- while dense correspondence is needed to build shape spaces, an expressive shape space provides a reduced dimensional space to regularize the search. 
We introduce \name, a method to solve both progressively. Starting from a small set of manually registered scans to bootstrap the process, we enrich the shape space and then use that to get new unregistered scans into correspondence automatically. 
The critical component of \name is a non-linear deformation model that captures details missed by the low-dimensional shape space, thus allowing progressive enrichment of the space. 

\let\thefootnote\relax\footnotetext{International Conference on 3D Vision, 2024}
\end{abstract}
\vspace{-.2in}    
\section{Introduction}
\label{sec:introduction}

Morphable models~\cite{blanz1999morphable,SMPL:2015,romero2017embodied} continue to strongly influence research towards human-centric workflows. This success is explained by the simple and versatile encoding of the underlying shape space, while providing interpretable handles for both shape and pose variations. The compact shape space has been extensively used for a variety of applications, including retexturing~\cite{garrido2014automatic}, shape editing~\cite{dale2011replacement}, pose and illumination manipulation~\cite{Wood_2021_ICCV}, animation~\cite{thies2015real}, avatar creation~\cite{hu2017avatar}, 
to name only a few. 

While morphable models are widely considered useful, creating them is surprisingly tedious. Theoretically, given a set of 3D shapes (e.g., scans of human bodies) with vertex-level correspondence,  morphable models can `simply' be built using linear (e.g., principal component analysis~(PCA)) or nonlinear (e.g., autoencoder~\cite{glass2022sanjeev,xu2020ghum}) dimensionality reduction methods. The hurdles lie first in getting scans of many subjects, with a wide coverage of body shape and pose variations, and second in establishing vertex-level dense correspondence across the scans. Given these challenges, not surprisingly, only very few high-quality morphable models (e.g., SMPL~\cite{SMPL:2015}, STAR~\cite{STAR:2020}, GHUM~\cite{xu2020ghum})  are publicly available.

The first hurdle has been significantly lowered with rapid advances~\cite{Bogo:ICCV:2015,dou2016fusion4d,dou20153d} in affordable, portable, fast, and robust (hardware) 3D scanning solutions (e.g., RGBD sensors, range scanners etc.). 
The second hurdle is algorithmic. The dominant approach to establish dense correspondence across the raw scans is to use non-rigid registration~\cite{anguelov2005scape} to align scans with a template (body) mesh. This works well when the input shapes have limited variations and are clean. Unfortunately, when shape variability is large (as among scans capturing representative variations across a population) or contains holes and noise, successful registration must rely on manual intervention or strong shape priors. 
Thus, either users have to annotate landmark correspondence across the scans, or provide shape priors to regularize the registration step.
Manual annotation is expensive and does not scale easily. Providing a shape prior is also tricky as generating one requires shapes in correspondence -- this leads to a chicken-and-egg problem.

We provide a solution that, starting from a small set of registered scans, alternates between building an underlying linear shape space and utilizing the current shape space model to automatically bring new (raw) scans into correspondence. At the core of our approach is a nonlinear deformation setup, expressed in the form of a neural network, i.e.,  Neural Jacobian Fields (NJF)~\cite{aigerman2022neural}, that helps to predict dense correspondence for scans \textit{close} to the currently modeled shape space. NJF is trained to add information beyond the current PCA space which is critical for registering new target scans, especially in the early stages where our linear shape space may yet not be sufficiently expressive. Once such correspondences are established, they enrich the shape space with additional scans. We repeat this process iteratively until all scans are brought into correspondence and a final shape space is achieved. We term this bootstrapping scheme Bootstrapped Linear Shape Space (\name ). In its current form, \name does not handle pose correctives.

We evaluate \name on the commonly used CEASAR dataset~\cite{CAESAR} and show that starting from only as few as $200$ manually registered scans, we can jointly learn a shape space and automatically bring additional  scans into correspondence. We evaluate the expressive power of the learned shape space on held out test scans and show that our model performs on par with models that require all scans to be registered manually. We also compare with standard non-registration methods and demonstrate that our method is more robust against noise and holes in the scans. Finally, we apply the methodology of \name in the context of face shape space construction to emphasize its generalizability. 

In summary, we make the following contributions: 
(i)~an on-the-fly PCA shape space and correspondence learning framework: starting from only 200 registered shapes, we progressively enrich the model with new shapes and eventually reach a shape space that is on par with the one trained with 3800 registrations directly; and
(ii)~a novel combination of linear PCA and non-linear Neural Jacobian Field~(NJF) deformation model that brings the target scans into better correspondence. 

\section{Related Work}
\label{sec:related}
\subsection{Non-rigid registration}
Registering two sets of raw scans (i.e., point clouds) is a long-standing problem \cite{bellekens2014survey,deng2022survey}, typically consisting of two steps: (i)~estimating correspondence between the source and the target scans; and 
(ii)~minimizing the distances between each correspondence pair to bring the source closer to the target. 
Since this work is concerned with 
human bodies that often deform non-rigidly, we review how correspondences are estimated in non-rigid registration of 3D human data.

\paragraph{Optimization-based (ICP).}
When the source and the target points are roughly aligned in the ambient 3D space, 
correspondences can be approximated by seeking nearest points. 
Following this intuition, existing methods \cite{groueix2018b,Hirshberg:ECCV:2012,huang2016bayesian,li2008global,Dyna:SIGGRAPH:2015} alternate between searching the closest point and deforming the source points,
which can be seen as non-rigid variants of the classical Iterative-Closest-Point (ICP) algorithm \cite{chen1992object,besl1992method}.
For fast convergence, such methods assume the two sets of points to be close enough, or  require an ``oracle guess" to initialize the correspondences. 

Furthermore, these methods often require additional regularization terms to avoid local minima, \eg, Laplacian \cite{sorkine2004laplacian} and ARAP \cite{sorkine2007rigid}. 
They impose extrinsic heuristics to constrain the deformation, which do not always apply to the target tasks.
In contrast, we employ the recently introduced Neural Jacobian Fields~(NJF)~\cite{aigerman2022neural} that implicitly learns an appropriate regularization in a data-driven manner.
We also use NJF in our method as it has been shown to better distribute error by having a global Poisson solve to integrate local gradient (\ie, Jacobian) information.

\paragraph{Learning-based shape matching.}
Global registration methods exist that match two human shapes without assuming they are close in 3D space. 
Instead of matching points in 3D space, they measure the similarity in a pre-defined feature space \cite{aubry2011wave,qi2017pointnet,sun2009hks,salti2014shot} and leverage machine-learning techniques to estimate correspondences \cite{boscaini2016,huang2017tracking,monti2017geometric,wei2016dense}, optionally refined with a global optimization \cite{azencot2019consistent,chen2015robust,ren2021discrete}.
The quality of these methods degrades significantly when the shapes are outside the distribution of the training data. 
More importantly, such methods do not yet handle noise in raw scans, and hence cannot be easily used in our setting. 
%
%

\subsection{3D Morphable Models for Humans}
A standard human body model has to account for pose and shape deformation. In this work, we are particularly interested in the latter -- the anthropometric variability across identities, and we focus the discussion on this aspect.

\begin{table}[b!]
\small
    \caption{Comparison with SMPL~\cite{SMPL:2015}, GHUM~\cite{xu2020ghum}, DenseRac~\cite{Xu_2019_ICCV}, and  STAR~\cite{STAR:2020} w.r.t.~the number of registrations used in training respective morphable models.}
\begin{tabular}{rccccc}
    \toprule
    \textbf{Method} & SMPL & GHUM & DenseRac & STAR & \name  \\
    \midrule  \midrule
    \# shape & 3800 & 64000 & 2500 & 15000 & 200 \\
    space & PCA & PCA/VAE & PCA & PCA & PCA \\
    \bottomrule
\end{tabular}
    \label{table:comparison_of_trianingmesh}
\end{table}

\paragraph{Parametric mesh.}
Representing human body parts as statistical shape models dates back to Cootes \etal~\cite{asm1995} in 2D and Blanz \& Vetter \cite{blanz1999morphable} in 3D.
The latter has become the de facto standard for modeling 3D human shapes \cite{egger20203d,tian2022hmrsurvey}, often called 3D morphable models (3DMM). 
The goal of a 3DMM is to adapt the template to each person by controlling the shape variations in a low-dimensional space.
In the context of whole body, a myriad of work \cite{anguelov2005scape,Joo2018_adam,SMPL:2015,STAR:2020,SUPR:2022,Pavlakos2019_smplifyx,xu2020ghum,Xu_2019_ICCV} has been proposed for this purpose and has led to rapid progress in monocular and multi-view human body reconstruction \cite{tian2022hmrsurvey}.

Learning such a parametric shape space, however, requires firstly, a large database of body scans, and more difficultly, bringing them into correspondence by registering a common template mesh to them. 
Most models above are trained with thousands or ten thousands of registrations to body scans in CAESAR \cite{CAESAR} and/or SizeUSA \cite{sizeusa}, 
curated with manual intervention for quality control (see Table~\ref{table:comparison_of_trianingmesh}). 
Another frequently overlooked challenge is that databases have each subject scanned in similar but not exactly the same pose (\eg, A pose in CAESAR) while the template is desired to be in one canonical pose (typically T pose).
To factor out the pose variation in the data, an un-posing process is performed to bring registrations to the canonical pose \cite{SMPL:2015}, which we refer to as ``canonical shapes" in the rest of the text.
Any artifact introduced in this step will be kept in the learned shape space.
Our formulation can take A-posed scans as input and output the canonical shapes in T-pose, requiring no un-posing before including them to training.

The most relevant work of Hirshberg \etal~\cite{Hirshberg:ECCV:2012} explores a ``semi-supervised'' setting of co-registration of multiple scans similar to ours. 
However, their approach differs from us in several aspects: 
(i) they aim to ``simultaneously'' learn a morphable model and bring scans into correspondence in one shot, without iterations. Consequently, it amounts to a big optimization problem where one has to provide good initialization (e.g., via manual landmarks) and carefully anneal the weights of each term, which may be easy to break. Due to the complexity of the optimization, this method can handle hundreds of scans whereas we can scale our iterative pipeline to thousands of scans; 
(ii) they rely on model-free registrations with a nearly isometric regularization term to capture information beyond the model space. While there is no publicly available code for us to perform an exact comparison, we compare to baselines where we employ a similar edge-preserving non-rigid registration approach and demonstrate superior performance.

%

\paragraph{Implicit surfaces.} A well-known limitation of meshes is that it is limited in handling deformations that require changes in the topology. 
Recent work \cite{deng2019neural,Mihajlovic:CVPR:2022,LEAP:CVPR:21,tiwari21neuralgif} explores representing human bodies with neural fields \cite{xie2022neuralsurvey}, which does not assume a consistent mesh topology.
They take the translations and rotations of body joints as input and estimate whether a query 3D point is inside the body or not. 
However, so far the effort has been devoted primarily in generalization of articulated poses, where large-scale motion capture datasets~\cite{ghorbani2021movi,AMASS:ICCV:2019} are used for training.

To help generalize to multiple subjects, it is encouraged to condition the networks on body shapes. 
However, in these methods, shape information is carried only in the locations of input joints, 
which is a very coarse anthropometric feature as two bodies can share the same joints but different surface shapes.
In this work, we consider explicit surface meshes in order to better capture details in human bodies.
\section{Method}
\label{sec:method}

\begin{figure}[b!]
    \centering
    \includegraphics[width=\columnwidth]{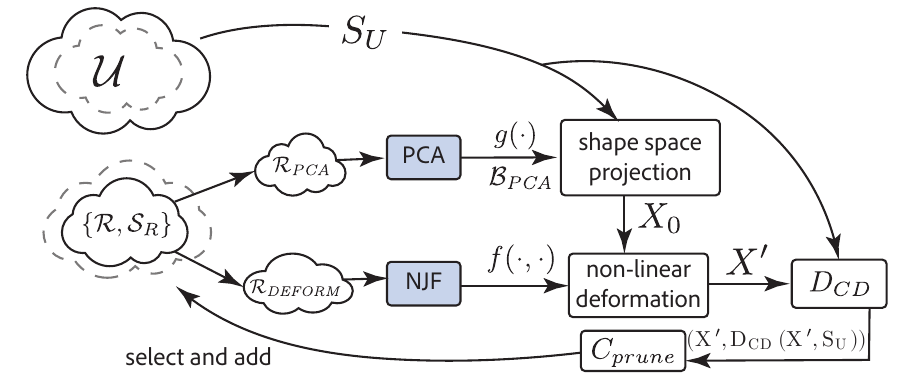}
    \caption{Given a sparse set of scans $\scanofregiset$, and their registrations $\regiset$ to a common template, we learn a linear shape space $\pcashapespace$ using $\regipca$ and train a non-linear NJF-based deformation model using $\reginjf$. Then, given a scan $\scan_U$ from a set of unregistered scans $\unregiset$, we project it to the PCA basis to obtain $\pcaproj$ and utilize NJF-based deformation to recover its registration to the template $\njfresult$ in the canonical pose. To enhance our shape space, we calculate the Chamfer Distance ($D_{CD}$) of registrations to target scans. We add all registrations where the distance falls within one standard deviation of the minimum distance to $\regipca$. We repeat this process to jointly register raw scans and enrich our shape space.}
    \label{fig:overview}
\end{figure}

\subsection{Overview}
Given a large set of raw scans \scans of varied human body shapes in roughly a similar pose (e.g., A-pose), our goal is to learn a shape space in a canonical but different pose (e.g., T-pose) that captures the variation of plausible body shapes. We learn a shape space with respect to a predefined shape template topology; in our setup, we use the SMPL~\cite{SMPL:2015} template, denoted as $\startemplate$ with $N$ vertices. 
We also assume having access to a small set of registered shapes $\regiset$ where a small set of scans \scanofregiset have been brought to the same topology as $\startemplate$ in the canonical pose via a manual non-rigid registration process to avoid any registration artifacts. Starting with $\regiset$ and $\startemplate$, we iteratively expand $\regiset$ with new shapes from the unregistered scans \unregiset that are automatically brought into correspondence with $\startemplate$ and learn an enhanced shape space. Note that we always have 
$\scans=\scanofregiset\cup\unregiset$. 

%
Our method works by deforming the template $\startemplate$ to closely match a new raw scan $\scan_U \in \unregiset$. This deformation model consists of two parts: (i)~a PCA-based shape space $\pcashapespace$ that provides a search space for shapes; and (ii)~a neural deformation model $f$ that maps shapes obtained by searching $\pcashapespace$ to targets that better capture the shape details of the raw scan. The two deformation models work in tandem to jointly register scans and yield correspondence with $\startemplate$, resulting in registered scans based on the current shape space. We then `close the loop' by selecting a few new registrations based on their distance to the scan and adding them to recompute a PCA basis $\pcashapespace$, thus enriching the shape space further. We repeat this process for multiple iterations, with each pass progressively learning a richer shape space and using it to register raw scans. Figure \ref{fig:overview} illustrates the pipeline of \name.

\begin{algorithm}[t!]
\footnotesize
\begin{algorithmic}[1]
\Procedure{\sc \name}{$\regiset, \scanofregiset, \unregiset, f, n$}

 \LeftComment{$\regiset$ = Registered Set, $\scanofregiset$ = Corresponding set of scans}
 \LeftComment{$\unregiset$ = Unregistered set of scans}
 \LeftComment{$f$ = Deformation Model, $n$ = Number of rounds}
 
 \For{Round $\in$ [1,$n$]:}
 \State {$\regipca,\reginjf \leftarrow \regiset$}
 \State {$\pcashapespace, g(\cdot) \leftarrow PCA(\regipca)$} \Comment{build shape sp.}

 \For{$\regi,\scanofregi \in$ ($\regiset$,$\scanofregiset$):} \Comment{build deform mod.}
    \State $\{ \alpha^\star_i \}, \theta^\star  = g(\scanofregi)$ \Comment{Fit using basis}
    \State $\pcaproj = \bar{S} + \sum_{i=1}^{k} \alpha_i^\star v_{s_i}$ \Comment{init.~canon.~shape}
    \State $\njfresult = f(\pcaproj,\scanofregi)$ \Comment{Register with NJF}
    \State $\lossvert = \lVert \njfresult - \regi \rVert ^ 2$ \Comment{Vertex Loss}
    \State $\lossjacob = \lVert J' - J \rVert ^ 2 $ \Comment{Jacobian Loss}
    \State $f(...;\gamma_i) = f(...;\gamma_{i-1})$ \Comment{Backprop}
 \EndFor

\State $C=\emptyset$ \Comment{New Candidate Registrations}
 \For{$\scan_U$ $\in$ $\unregiset$:}
    \State $\{ \alpha^\star_i \}, \theta^\star = g(\scan_U)$
    \State $\pcaproj = \bar{S} + \sum_{i=1}^{k} \alpha_i^\star v_{s_i}$
    \State $\njfresult = f(\pcaproj,\scan_U)$
    \State $C \leftarrow C \cup \njfresult$
 \EndFor
 \State $D$ $\leftarrow$ $\text{ChamferDist}(C, U)$
 \State $\text{th}$ $\leftarrow$ $\min(D) + \sigma(D)$
  \State $C_\mathit{prune} \leftarrow \{c | c \in C$, $\text{ChamferDist}(c, U) < \text{th} \}$ 
 \State $\regiset \leftarrow \regiset \cup C_\mathit{prune}$ \Comment{Update Basis}
 \EndFor
 \State return \regiset
 \EndProcedure
 \end{algorithmic}
\caption{}
\label{alg}
\end{algorithm}

\subsection{PCA-based Shape Space}
\label{sec:pca}
We use a subset of the shapes in $\regiset$, $\regipca$ to compute a PCA basis in a similar fashion to classical works like SMPL~\cite{SMPL:2015} and STAR~\cite{STAR:2020}. Note that we use only a subset of the \regiset and save the rest for the data-driven deformation model (see Section~\ref{sec:njf}). 
Our shape space $\pcashapespace$, similar to others, is composed of a pose-corrective deformation basis allowing for pose-conditioned deformations and a shape basis that enables body-shape deformations. 
In our work, since we are primarily interested in learning a space of body shapes, we borrow the pose corrective directly from SMPL, which is denoted as $B_P(\pose): \mathbb{R}^{\| pose \|} \rightarrow \mathbb{R}^{3N}$ as well as a rigged skeleton to pose $\startemplate$, where $\| pose \| = 24 \times 3$, corresponding to 3 axis-angles for each of the 24 joints. 
We represent the shape basis $\pcashapespace$ with $k$ shape eigenvectors $\pcashapespace := \{v_{s_i}\}$, where $k$ is selected such that the shape variation in the dataset is explained using the $k$ basis vectors. 
In this computed space, we define a new shape $\canonical$  in any particular pose $\theta$ as, 
\begin{align*} 
\label{eq:canonical}
\canonical( \{\alpha_i\}, \theta) & :=  \bar{S}  + \sum_{i=1}^{k} \alpha_i v_{s_i} + B_P(\pose) \\
\posed( \{\alpha_i\}, \theta) &:= \mathcal{W}\left(\canonical(\{\alpha_i\}, \theta), \mathcal{J}, {\theta}, {W}_s\right),
\end{align*}
where $\bar{S}$ is the mean shape, $\mathcal{J}$ is the joint regressor that provides the joint locations given the vertex positions in the shape, ${W}_s$ is a fixed set of skinning weights, and finally, $\mathcal{W}$ is the skinning function as defined in \cite{SMPL:2015}. 
Now, given a target scan $S$ and a current set of shape basis vectors $v_{s_i}$, we optimize for the pose parameters and shape coefficients:
\begin{eqnarray}
\label{eq:optimize}
g(\scan_U)&:= & ( \{ \alpha^\star_i \}, \theta^\star  ) \nonumber \\ &= & \arg\min_{\{\alpha_i\},\pose} D_{CD}(\mathcal{W}(\canonical, \mathcal{J}, \pose, {W}_s), \scan_U),
\end{eqnarray}
where $D_{CD}$ is the Chamfer Distance and $\scan_U$ is an unregistered raw scan.

We optimize Eq.~\ref{eq:optimize} to find the shape in $\pcashapespace$ that best matches the scan $\scan_U$ while also optimizing for the pose parameters $\pose$. In other words, the function $g$ ``projects" the raw scan $\scan_U$ onto the shape space $\pcashapespace$. After optimization, we obtain the canonical shape that corresponds to the scan as $\pcaproj := \bar{S} + \sum_{i=1}^{k} \alpha_i^\star v_{s_i}$.
Note that due to the limited expressivity of the linear basis, $\pcaproj$ may not accurately represent $\scan_U$. 
We now seek a deformation model that can further enrich $\pcaproj$ with the details from $\scan_U$. 
This shape space optimization also provides a good initial point to seed our subsequent nonlinear deformation model, as explained next.

\subsection{Neural Deformation with NJF}
\label{sec:njf}
In our work, a nonlinear deformation is simply an assignment of new 3D positions to the vertices of the given (template) mesh. We adopt Neural Jacobian Fields (NJF) \cite{aigerman2022neural} as our nonlinear deformation model $f$. NJF trains an MLP to map triangles on a source mesh to a corresponding deformed triangle on a target mesh using only local information. The key step is to have this training receive gradients through a differentiable global Poisson Solve layer to then directly predict the positions of the vertices. 

We consider another subset of shapes in $\regiset$, $\reginjf$ where $\reginjf \subset \regiset \setminus \regipca $, and their corresponding raw scans 
to train NJF. For each shape $\regi \in \reginjf$, we first optimize for parameters $g(\scanofregi)$ where $\scanofregi$ is the raw scan corresponding to $\regi$, giving us a shape space projection $\pcaproj$. 
We then train NJF to map the canonical $\pcaproj$ to the canonical $\regi$, \textit{conditioned} on the scan $\scanofregi$ that can be in any pose. 
Essentially, we ask our deformation model $f$ to deform the result of our (current) shape space projection $\pcaproj$ to the target registration $\regi$ that contains richer details.
The deformation function $f$ is conditioned on the raw scan representing the target, and can fix any residues not covered by the optimization step. Specifically, we train $f$ with per-vertex $L_2$ loss,
\begin{equation}
\label{eq:njfloss}
    \lossvert := \lVert f(\pcaproj,\scanofregi; \gamma) - \regi \rVert ^ 2,
\end{equation}
where $\gamma$ represents learnable parameters, and a per-triangle Jacobian loss $\lossjacob$ which supervises the ground-truth Jacobians $J$~(see \cite{aigerman2022neural}), with our total training signal being,
\begin{equation}
\losstotal = 10 \cdot \lossvert + \lossjacob.
\end{equation}
We slightly abuse notation in Equation \ref{eq:njfloss} -- in practice,  $\pcaproj$ and $\scanofregi$ are represented as features and not by vertex locations themselves. Specifically, we use Pointnet encodings \cite{DBLP:journals/corr/QiSMG16} of both shapes. Details of the network architectures and the features used are  in the supplemental.

With our initial shape space defined by $\regipca$ and a nonlinear deformation model trained with $\reginjf$, we can now use these in tandem to register new scans.

\subsection{Closing the Loop}
\label{sec:loop}
For each unregistered scan $\scan_U \in \unregiset$, we first fit the template $\startemplate$ to it by optimizing parameters $g(\scan_U)$ in Equation~\ref{eq:optimize}, to obtain the canonical pose $\pcaproj$. 
We then use the trained NJF to predict the final registration as $f(\pcaproj,\scan_U;\gamma) \rightarrow \njfresult $. 
The model \njfresult is then ``posed'' to match the pose of $\scan_U$ by using the optimized pose parameters $\pose^\star$. 
We compute the Chamfer Distance to their corresponding scans, and if the error falls within one standard deviation from the minimum error, we augment $\regipca$ with the new shapes \njfresult in T-pose.
In the next iteration, the shape space will be updated by computing PCA with the augmented set $\regipca$. The updated basis also provides new initial states for training our deformation model $f$. 
Note that we do not add the PCA projections \pcaproj to \regipca as it does not carry ``fresh'' information like \njfresult.

We repeat the steps of constructing a PCA basis, learning an NJF based deformation model, and registering new scans for several rounds, with each round expanding the shape space (see Algorithm \ref{alg}).


\begin{table*}[t!]
\centering
    \caption{Evaluating ours against alternatives. 
    (i)~Learning a one-time static shape space from 400 available registrations provides an upper bound;  (ii) and (iii) provide baselines replacing our non-linear deformation model with classical non-rigid registration. Errors are in cm.}
\begin{tabular}{r rccccc}
    \toprule
    \textbf{Method} & $|\regieval|$ & initial $|\regipca|$ & $|\reginjf|$ & regularizations & \# shapes $\in \unregiset$ & v2v ($\downarrow$) \\
    \midrule  \midrule
 (i) \baselinefullpcanjf    & 29    & 400 & 400    & \xmark & \xmark & {\em 0.87} \\
    \midrule  
    (ii)~\baselineC - PCA + non-rigid     & 229   & 100 & \xmark & small $\|\Delta v\|$ & \numunregiscanVtwo & 3.11 \\
    (iii)~\baselineD - PCA + non-rigid    & 229   & 100 & \xmark & edge-preserving & \numunregiscanVtwo & 3.26 \\
    \oursvtwopcaonly       & 229   & 100 & 100    & \xmark & \numunregiscanVtwo & 1.31 \\
    \oursvtwo (PCA+NJF)       & 229   & 100 & 100    & \xmark & \numunregiscanVtwo & \textbf{0.90} \\
    \bottomrule
\end{tabular}
    \label{table:baselines}
\end{table*}

\begin{figure*}[t!]
 \centering
\includegraphics[width=\linewidth]{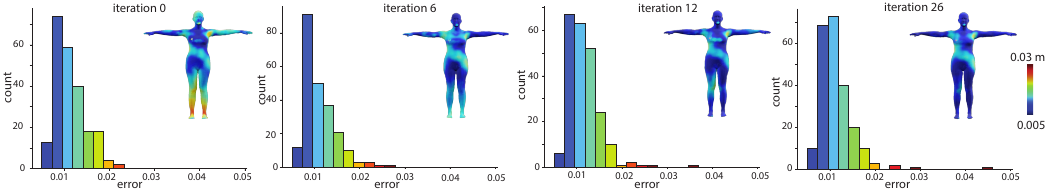}
\caption{We show the histogram of the v2v error of the scans in our test set at different iterations of our method. We also color code the per-vertex error for an example scan. As our method progresses, the error decreases, and we observe a slight left shift in the histogram as the shape space improves. Insets show residue error on one scan over iterations.}
 \label{fig:residuehistogram}
\end{figure*}

\section{Experiments}

\subsection{Dataset and Protocols}

\textbf{Dataset.} We picked \totalartistshape scans from the CAESAR dataset and had them registered by a professional artist, \ie,  $|\regiset|=429$. Note that the artist took 40-60 min per scan using a combination of landmark point specification, running nonrigid ICP, and then manually fine-tuning dense correspondence correction/specification (\eg, around fingers, armpit, etc.), costing around \$25 for each scan. 
We consider these artist-registered meshes  as Ground Truth for evaluation, training, and, in the case of some baselines, as targets. 
Specifically, we first sample two mutually exclusive sets \regipca and \reginjf from \regiset to train the initial PCA space and NJF respectively, where $|\regipca|=100$ and $|\reginjf|=100$. We use all the remaining 229 registrations as \regieval for evaluation purposes unless noted otherwise. 



Since the original CAESAR dataset consists of around 4000 scans, 429 of which we have registrations from the artist, we consider the rest 3.5k scans as unregistered scans \unregiset.
In Algorithm \ref{alg} Ln.~14-18, scans in \unregiset are brought to correspondence, added to \regipca, and contribute to the new shape space \pcashapespace, whereas \reginjf is kept fixed to the initial 100 artist registrations. 
Throughout the experiment, we always use $k$=11 basis for any PCA-based shape space, denoted as PCA (see supplemental for tests with other $k$ values).
Note that despite \reginjf being fixed, in each round, since the basis of the shape space \pcashapespace changes, \pcaproj changes, and consequently, the amount of details that NJF needs to compensate also changes. 
Hence, our NJF is rebuilt in each iteration (Algorithm \ref{alg} Ln.~5-11).
At test time, since \name consists of a PCA shape space and an NJF network, 
whenever we evaluate only the learned PCA space, we denote it as ``PCA only'' to distinguish from running the full pipeline.

\textbf{Metrics.} For each registration in \regieval, we take the corresponding raw scan and obtain a registered shape using either our method or the baselines below. 
We measure the vertex-to-vertex (v2v) distance between the ground truth and the estimated canonical shapes, using the artist-annotated scan-to-template correspondences. When comparing shape spaces with different topology, such as GHUM, we start by non-linearly registering the target and our mean body shape models in the same pose. Then, we use barycentric coordinates on the corresponding face to map each point on our template to the target body model. Additionally, we calculate the vertex-to-plane error (v2p), which does not require meshes to have the same topology.

%
(i)~\textbf{\baselineB}:  We spare 29 registered shapes for evaluation and use all the remaining 400 annotated meshes to train a PCA model. We further train an NJF with the same 400 scans to add the missing details not covered by the PCA model, denoted as ``\baselinefullpcanjf''.
This baseline represents scenarios where one trains the model \emph{in one go} with all available registrations, without any bootstrapping schemes that leverage the unregistered scans. Hence, this can be seen as an {\em upper bound}.

\textbf{Baselines.}
We consider several baselines, where NJF is replaced in our pipeline with classical non-rigid registration methods.
Given an unregistered scan $\scan_U$, we first obtain the projection in the PCA space \pcaproj and then optimize the location of each vertex on \pcaproj, such that when posed with $\pose^\star$, the shape yields low Chamfer Distance to the scan $\scan_U$.
This ``free-form deformation'' scheme can fall into local minimum easily even if we provide \pcaproj as close initialization. Therefore we define new baselines where we constrain it with standard regularization terms: (ii)~\textbf{\baselineC}: vertices should not be deviating too far from the canonical shapes \pcaproj, \ie, $\|\Delta v\|$ should be small favoring smooth surfaces; (iii)~\textbf{\baselineD}:  deformation should preserve edge lengths, \ie, favor near-isometric deformations.


Finally, we also compare to existing shape spaces including SMPL~\cite{SMPL:2015}, STAR~\cite{STAR:2020}, and GHUM~\cite{xu2020ghum}.


\subsection{Results and Discussions}

\textbf{Progressive improvement of the shape space.} First, we illustrate how our shape space is progressively improved, \ie, becomes more expressive. In Figure \ref{fig:residuehistogram}, we show the histograms of the v2v residue error at different iterations of our method as it consumes new scans from $\unregiset$ and visualize the error as heat maps in each inset image.
One can observe that the error gets reduced as we have more rounds, \ie, the correspondence quality improves progressively.

Further, we compare with a method that simultaneously burns all available 400 scans into model training -- \baselineB.
As shown in Table \ref{table:baselines} upper part, it attains the lowest v2v error of 0.87cm on a smaller evaluation set of 29.
In the bottom part, we can observe that, despite being trained initially with only 100 registrations, \name yields v2v error that is on par with the upper bound by automatically extending the set of registered scans to 800. 

\textbf{Effect of NJF as a non-linear registration module.} In \textsc{Baseline}1-2, we run Algo.~\ref{alg} but estimate detailed shapes \njfresult at Ln.~16~by non-rigid registration methods instead of NJF $f$. 
We then evaluate how well the resulting PCA shape space explains the test set \regieval and compare with our PCA shape space.
In Table \ref{table:baselines} bottom part, we observe that after consuming \numunregiscanVtwo unregistered scans, 
our shape space explains scans in \regieval with 1.31cm v2v error (PCA only), while NJF further reduces it to 0.90cm.
Consuming the same amount of scans, shape spaces enriched with \njfresult from non-rigid registration yield errors of 3.11cm and 3.26cm, respectively. 
This suggests using a data-driven NJF in the loop recovers better correspondence than optimization-based registration methods, and when included in \regipca, it leads to a new PCA space \pcashapespace with richer information.

Figure~\ref{fig:residuehistogram} shows how the model improves over iterations/rounds (from left to right). While NJF helps add lost details, it is still limited by the sparse set of training pairs (100 in our case) and may not recover all the details.

\begin{figure}[b!]
  \centering
    \includegraphics[width=\linewidth]{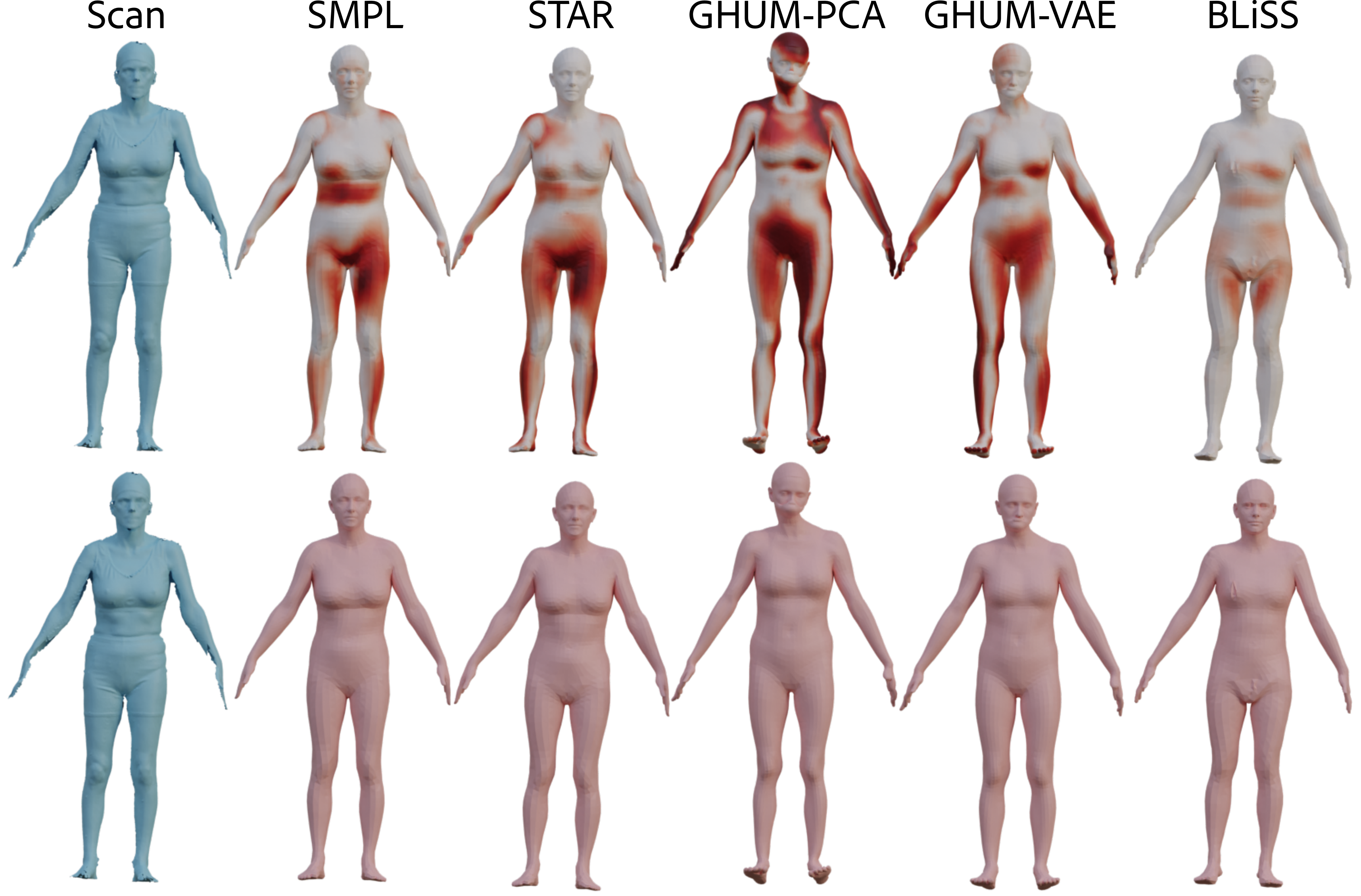}
   \caption{For a given raw scan, we register each body model by predicting pose and body shape parameters. (Top) Each result is color coded based on the v2p error in meters w.r.t.~the ground truth registration provided by the artist.}
   \label{fig:oursvssmpl}
\end{figure}

\textbf{Comparison to other shape spaces.} We compare \name with the following existing shape spaces: (i) the classical SMPL~\cite{SMPL:2015} shape space trained with the registrations of 3800 CAESAR scans, (ii) its follow-up STAR~\cite{STAR:2020} which uses additional 11000 registrations of the SizeUSA dataset~\cite{sizeusa}, (iii) GHUM which uses registrations for an additional proprietary dataset of 64000 scans (where a majority consists of body, hand, and, facial pose variations) along with CAESAR. GHUM presents a linear shape space as well as a VAE-based non-linear shape space,  both of which we include in our comparisons.
For each corresponding scan in \regieval, we optimize for the pose and shape parameters of each body model and report both the v2v and v2p errors in Table \ref{tab:vsSMPL}. We also show qualitative comparison in Figure~\ref{fig:oursvssmpl} where we color code each optimized body model using the v2p error with respect to the ground truth artist provided registration.

\begin{table}[h!]
\centering
\footnotesize
\caption{Ours, after absorbing 800 shapes from CEASER, outperformed SMPL, STAR, and GHUM even though we only used 200 registered scans, compared to their much larger number of scans. }
\begin{tabular}{rccc}
\toprule
\textbf{Method} & \# shapes $\in \regiset$ & v2v ($\downarrow$) (cm) & v2p ($\downarrow$) (cm) \\
\midrule  \midrule
SMPL (PCA)  & 3800            &   1.72 & 0.62 \\
STAR (PCA)  & 15000            &   2.15 & \textbf{0.58} \\
GHUM-PCA   & 64000            &  6.74  & 3.01 \\
GHUM-VAE   & 64000            &  5.89  & 2.63 \\
\oursvtwo           & 200 &  \textbf{0.90} & 0.65 \\
\bottomrule
\end{tabular}
\label{tab:vsSMPL}
\end{table}

We observe that \name yields consistently lower v2v error than other shape spaces. We use 11 PCA components for SMPL, STAR, and \name while all the PCA components for GHUM.
Our Ground Truth are artist-annotated registrations, which can still potentially contain errors which might have an effect on this gap. Nevertheless, our method performs on par with SMPL and STAR based on the v2p error and better than GHUM.
Hence, the primary observation in Table \ref{tab:vsSMPL} is that, despite starting from only a small amount of registrations (100+100), \name yields on-par expressivity compared to a model trained with an order of magnitude more registrations. 
We attribute this to the novel combination of linear PCA and non-linear NJF deformation model, as well as the progressive scheme leveraging such a hybrid deformation model for better correspondence. 

\begin{table}[b!]
\centering
 \footnotesize
    \caption{We use Farthest Sampling to gather 500 shapes from each space. To determine the similarity between the different spaces, we calculate the distance between each shape in one space and its closest counterpart in all other spaces, including off-diagonal entries. We then report the average distance (in centimeters)  for each possible pairing of spaces in both directions. Low values for (A, B) and (B, A) suggest the two spaces are similar.
    We compute the diversity of samples inside each space for diagonal entries, with higher values indicating more diversity. 
}
\begin{tabular}{rccccc}
    \toprule
    \textbf{Space}& Ours & GHUM & STAR & SMPL \\
    \midrule  \midrule
    Ours & \textcolor{peach}{(4.10)} & 3.57 & 1.38 & 1.46 \\
    GHUM & 4.03 & \textcolor{peach}{(4.48)} & 3.65 & 3.71 \\
    STAR & 1.79 & 3.74 & \textcolor{peach}{(3.96)}  & 1.37 \\
    SMPL & 1.90 & 3.75 & 1.36 & \textcolor{peach}{(4.14)} \\
    \bottomrule
\end{tabular}
\vspace{-.1in}
    \label{table:baseline-div}
\end{table}

\begin{figure}[h!]
    \centering
    \includegraphics[width=\columnwidth]{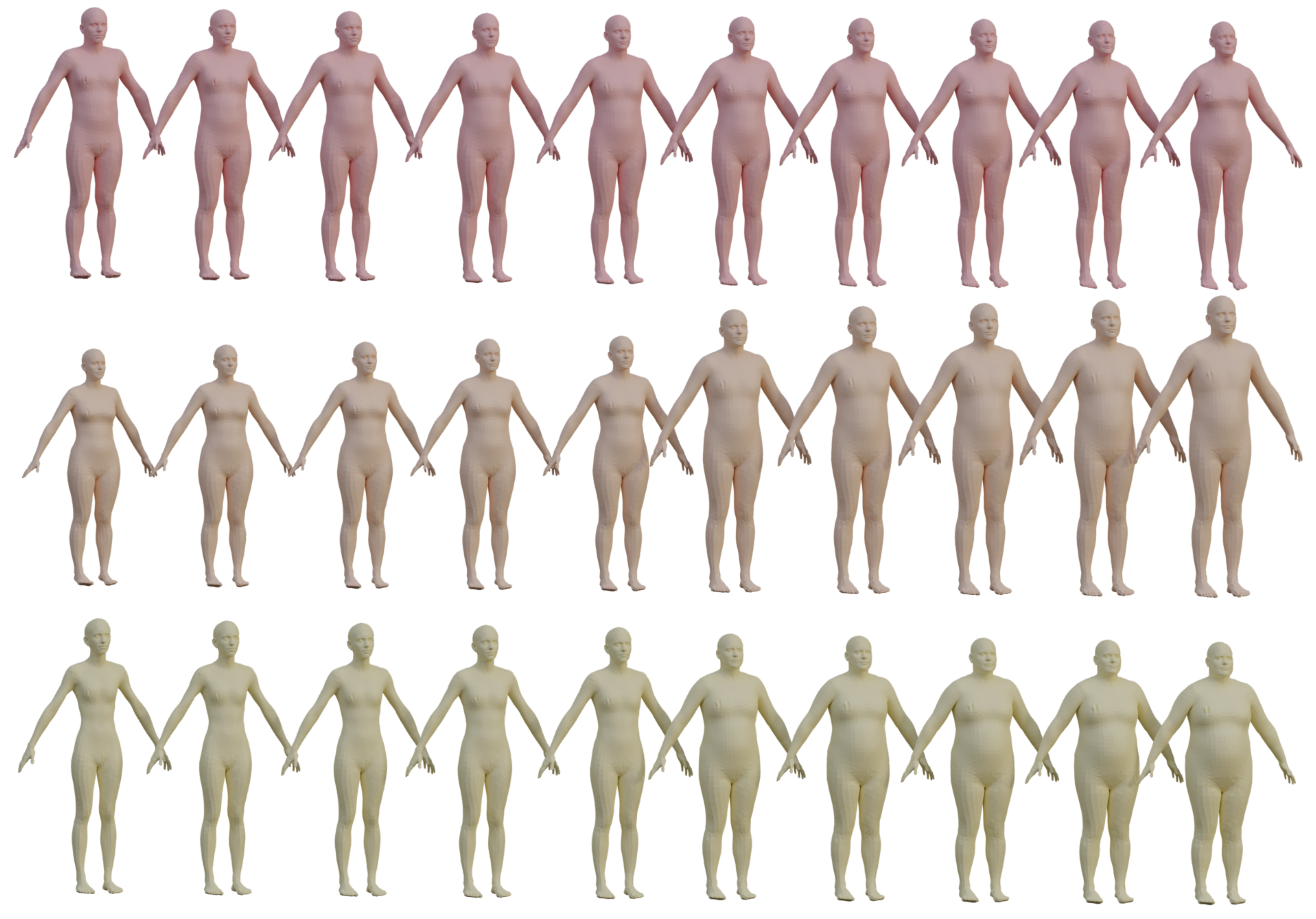}
    \caption{We show shapes along the top three principal directions in different rows, and observe variations in gender, height, and weight along the respective PCA  modes.}
    \label{fig:shapeSpace_pca}
\end{figure}

\textbf{Diversity of body shape spaces.}
We qualitatively show the diversity of body shapes represented by our shape space by randomly sampling our final PCA space using farthest sampling in terms of vertex differences. Sampled shapes are shown in Figure \ref{fig:teaser}. In Figure \ref{fig:shapeSpace_pca}, we show shape variations captured by our shape space's top three PCA modes.

\begin{figure*}[t!]
 \centering
\includegraphics[width=\linewidth]{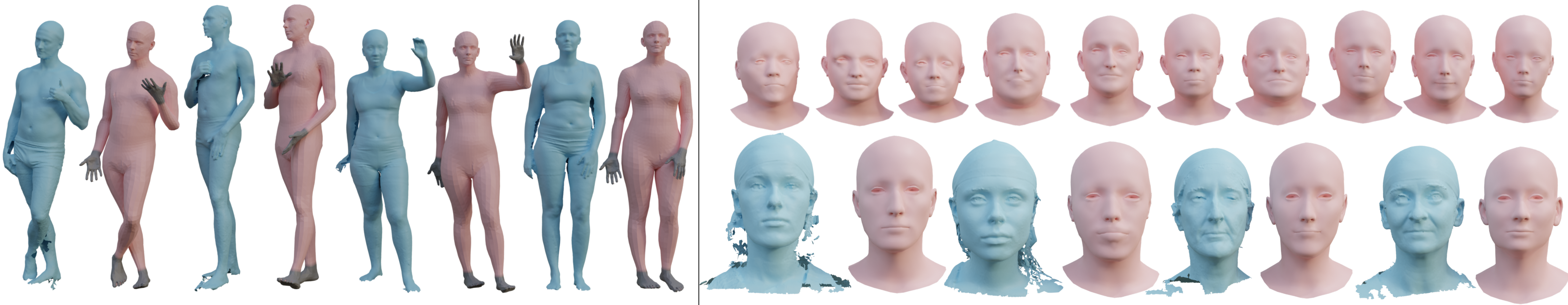}
\caption{\textit{Left:} Registration (pink) of noisy scans (blue) with our final shape space. Since our model does not capture finger-level details, after optimization, the joints corresponding to the greyed-out regions are reset to default poses. \textit{Right}: We show sampled faces from our final face-shape space after growing it from $20 \rightarrow 800$ shapes. We observe a variety of face changes in the cheek and nose regions. (Bottom) We take the test scans from the COMA dataset (in blue) and register them in our final face-shape space, which is shown in pink.}
 \label{fig:bodyface}
\end{figure*}

In order to compare the diversity of our and existing shape spaces, we sample 500 body shapes in each shape space by furthest point sampling. For each sampled body shape, we compute is nearest sample within the same shape space by measuring the v2v error. We report the average of such pairwise sample error along the diagonal of Table~\ref{table:baseline-div} where higher pairwise distance means a diverse shape space. As shown quantitatively, the diversity of our shape space is on par with existing shape spaces. We also compute each sample in one body shape space's nearest sample in all other shape spaces. For each shape space in each row, we compute the pairwise sample distance for each shape space in each column. Smaller numbers indicate the similarity of shape spaces. We observe that our shape space is closer to the SMPL and STAR shape spaces.

\textbf{Number of PCA components.} While we use k=11 basis consistently for all PCA-bsaed methods, we also analyze the effect of using more bases with k=30, 50, 100. Using higher number of basis increases the expressivity of PCA, but empirically we observe very little change in our metrics – to the order of $10^{-5}$ – by considering more PCA components, and thus stick to using just k=11 in all experiments. 

\textbf{Application.}
A typical application of a body shape space is to predict a given raw scan's shape parameters. We demonstrate the use of \name shape space in such an application in Figures~\ref{fig:bodyface} and \ref{fig:teaser} (right) . Since our work focuses on capturing body shape variation, we optimize for pose in SMPL's pose space. For each raw scan we use 9 manually annotated landmarks to estimate the initial pose, then we estimate the body pose (with SMPL) and shape parameters in our shape space. We observe that our space accurately estimates the body shape despite the scans being noisy.


\textbf{Face registration.}
To demonstrate the generalization of our approach, we sample 20 faces from FLAME \cite{FLAME:SiggraphAsia2017} to create an initial face shape space. We then iteratively register faces from the COMA \cite{COMA:ECCV18} dataset. Note that we use the NJF module trained on full-body human scans. Registrations shown in Fig.~\ref{fig:bodyface}. We do \textit{not} assume paired data for this task and instead exploit NJF's invariance to topology and use the pre-trained NJF to deform the faces. NJF was trained on centroids and Wave Kernel Signature as input features; these features are computed for the linearly registered face scans and fed to NJF to add further details. We iteratively train \name until we add  800 face-shapes to ours.

\section{Conclusions}

We have presented a method that takes in a small set of artist-annotated scans along with a much larger corpus of unregistered scans, and jointly learns a (linear) shape space while progressively bringing the unregistered scans into correspondence. At the core of our approach is a novel formulation that continuously refines the underlying shape space and a learned nonlinear module that automatically registers models `close' to the current shape space. We demonstrate that our approach trained only with 200 registered scans, can produce competitive performance compared to established shape space models,  trained using thousands of registered scans. We also presented similar results for face datasets. 

One limitation of our method is that it does not capture pose corrective shape space. To address this, we plan to use the non-linear deformation module in an iterative fashion to learn a pose corrective shape space from unregistered scans. We may also explore the use of nonlinear shape spaces, such as AE or VAE, but it poses a challenge of growing robustly on limited data, especially in the initial rounds of the approach. It is worth noting that our method cannot handle hands in complex poses due to the lack of finger articulation in SMPL's pose space. 

It is advisable to minimize the use of shape regularizers like ARAPReg~\cite{arapReg:21}, data-efficient shape VAE~\cite{glass2022sanjeev} as they may limit the flexibility of the learned shape spaces. Instead, we can adopt an alternate approach by directly selecting models and asking artists to annotate them, maximizing fresh information beyond fully self-supervised shape space building.

\paragraph{Acknowledgement} This project has received funding from the UCL AI Center, gifts from Adobe Research, and EU Marie Skłodowska-Curie grant agreement No 956585.
{
    \small
    \bibliographystyle{ieeenat_fullname}
    \bibliography{main}
}
\clearpage
\setcounter{page}{1}

\twocolumn[{%
\renewcommand\twocolumn[1][]{#1}%
\maketitlesupplementary
\maketitle
\thispagestyle{empty}
     \includegraphics[width=\linewidth]{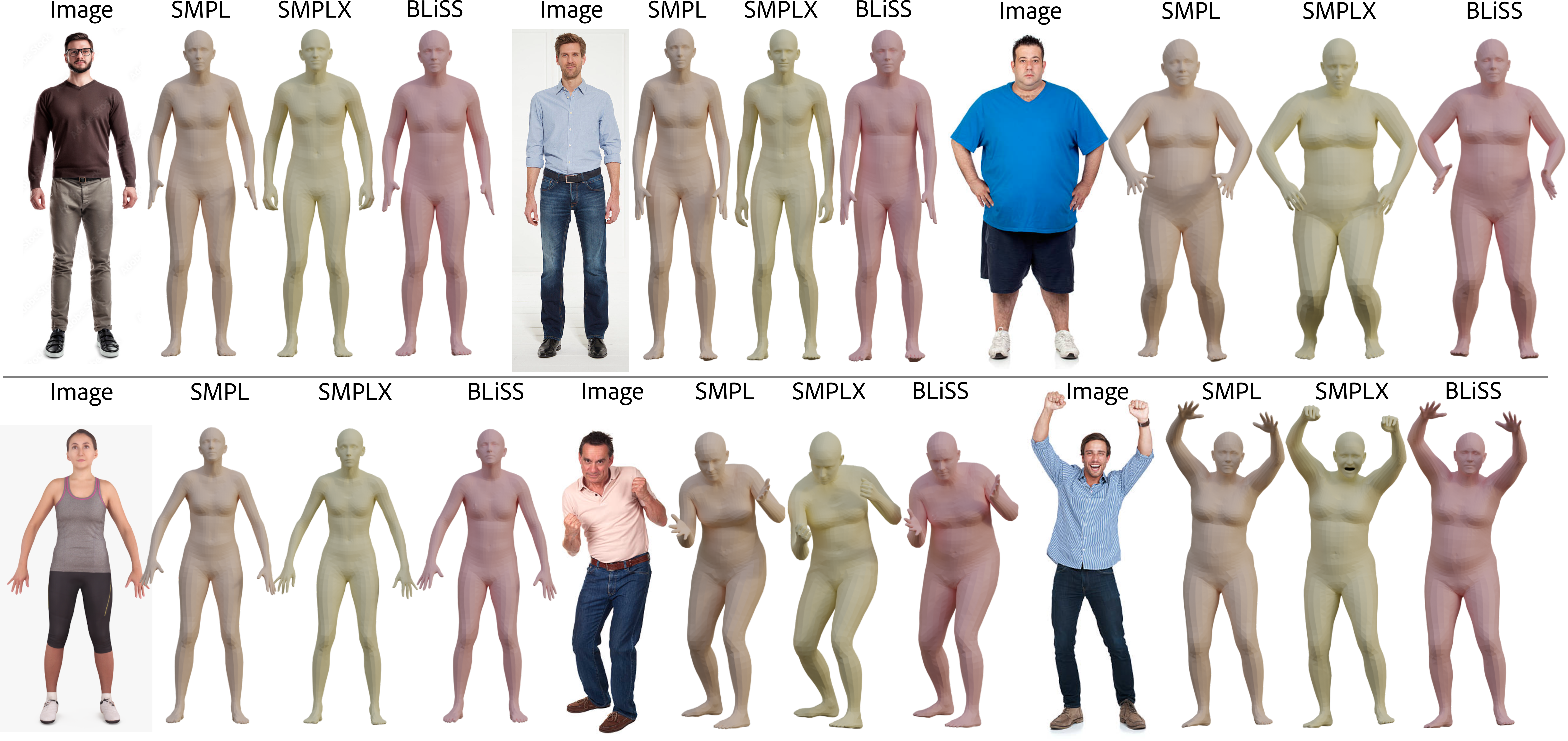}
     \captionof{figure}{Here we use the trained SMPLify-X \cite{Pavlakos2019_smplifyx} model to estimate the shape from a single image. For \name, we plugin our shape space as a drop-in replacement for SMPL's space, while using SMPL's pose space.}
    \label{fig:smplify}
    \vspace{+0.50em}
}]
\section{Shape Estimation from Single Image}
\vspace{-1mm}
We additionally compare our shape space with SMPL's by plugging them into an 
optimization-based body reconstruction framework SMPLify-X \cite{Pavlakos2019_smplifyx}. Since \name doesn't handle pose, we use SMPL's pose space. SMPLify-X then searches the given space to fit the shape to the given image. We observe in Figure \ref{fig:smplify}, that \name's estimated shape is comparable with that of SMPL and SMPL-X, despite having started from only 200 given registrations. Note that unlike SMPL-X, SMPL (and hence \name) does not articulate hand poses.
\vspace{-2mm}

\section{Iterative improvements}
\begin{figure}[h!]
    \centering
    \includegraphics[width=.8\columnwidth]{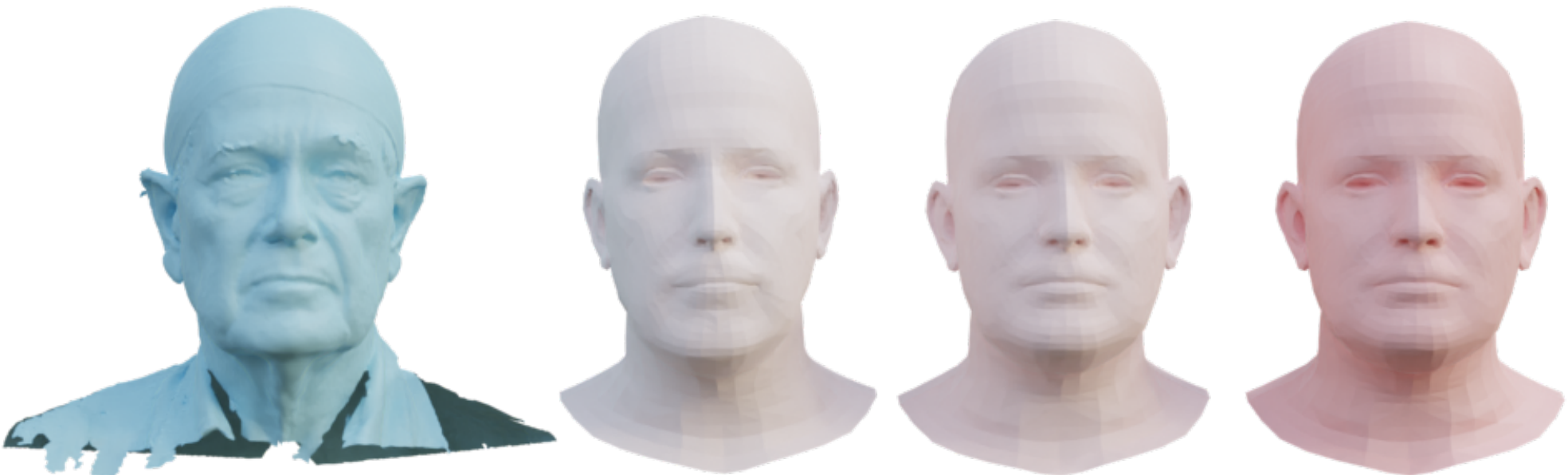}
    \caption{{\textbf{Iterative Shape Corrections:} Lightly colored faces in the middle are our registrations in earlier iterations of the space, and the pink-colored face on the right is our registration after five rounds of \name. As the rounds progress, registrations in later rounds more accurately capture the scan (left, in Blue), as observed by the broadening of the nose and jawline. }}
    \label{fig:facerounds}
\end{figure}

\vspace{-2mm}
Registrations in the initial rounds of our shape-space are often coarse, lacking details, as shown in the two pale colored faces in the middle in Figure 2 \ref{fig:facerounds}. As the space is more densely populated, later registration more closesly capture finer details of the given scan.

\vspace{-2mm}
\section{Hand Registration - MANO}
\begin{figure}[h!]
    \centering
    \includegraphics[width=\columnwidth]{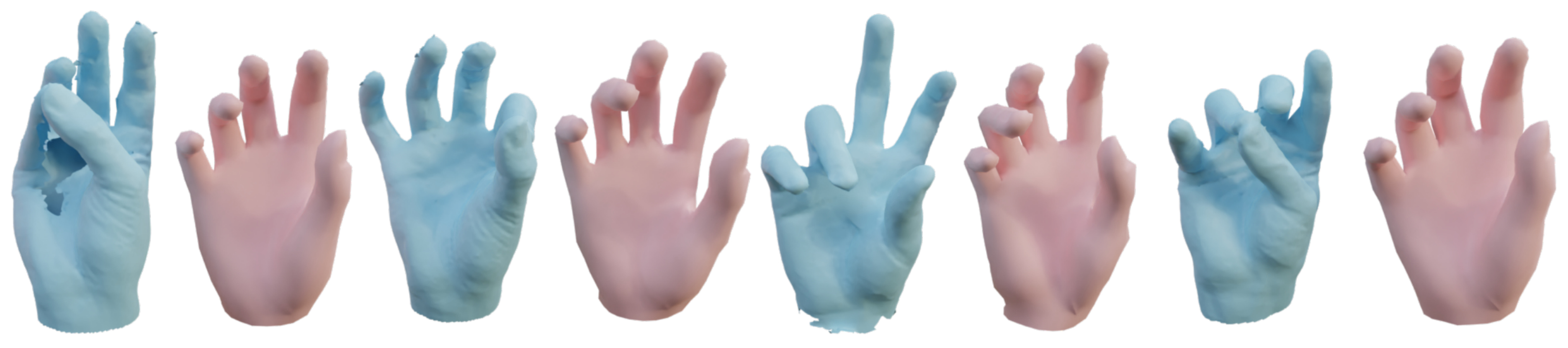}
    \caption{{\textbf{Registering Hand scans:} We use our final hand-shape space to register hand scans (blue); registrations in canonical pose (i.e., default) shown in pink.}}
    \label{fig:hand}
    \vspace{-.1in}
\end{figure}

\vspace{-2mm}
We sample 20 hands from MANO to create an initial shape space. We then iteratively register scans from the MANO dataset using the pre-trained NJF \textit{without} assuming paired data. Since NJF is a topologically invariant mapping module, we are able to train the network for full body registrations, and use the trained MLP to register out-of-domain scans. Registrations are shown in Figure~\ref{fig:hand}.
\vspace{-4mm}

\section{Implementation Details}
\subsection{Features for Neural Jacobian Field}
\vspace{-2mm}
At the core of NJF is a Multi-layer Perceptron (MLP) that processes the input features on each triangle of the given mesh to produce a per-triangle Jacobian, which is used in a differentiable Poisson solve to compute the deformed vertex positions. We use NJF to deform our PCA projection $X_o$ conditioned on the scan $S_X$, similarly to the shape morphing experiments proposed in the original paper~\cite{aigerman2022neural}. In our tests, we used the following features: 
\begin{itemize}
    \item Since $S_X$ is a raw scan, the PointNet encoding of its coordinates represents it. We use both the global encoding of the scan and its per-point features from PointNet. Since the scan and $X_o$ are \textit{not} in correspondence, we choose features of those points that are closest to a point on $X_o$. We note that despite $X_o$ and $S_X$ having different poses, the nearest neighbor feature look-up provides an indication to the MLP of the kind of shape transformation that is required.
    \item We associate with each triangle of $X_o$, the PointNet encoding of $S_X$ and $Ss_X$'s points as obtained above, and of $X_o$ itself. Specifically for $X_o$ as in the original work, we encode each triangle's centroids, normals and top 50 Wave Kernel Signatures \cite{aubry2011wave}. In addition, we also pass the Jacobians of $X_o$ itself as we observe that it significantly improves the mapping to target. Note that $S_X$ and $X_o$ are processed via different PointNets as their input features are different
    
\end{itemize}
We pass the above features to a 4-layer MLP, with each hidden layer being 128 wide and activated by ReLU. The final Linear layer produces a 9-dimensional vector for each triangle (as a Jacobian is a $3\times 3$ matrix). Both PointNets - one for $X_o$ and one for $S_X$ - and the MLP are trained jointly to produce the mapping from $X_o$ to the desired shape.

\vspace{-1mm}
\subsection{Summary of the Use of CAESAR data}
\vspace{-1mm}
We have registration information for 429 of all  4000 CAESAR scans. 
We use 100 of them for training the initial PCA shape space and another 100 for NJF, 
withholding the rest 229 for evaluation purposes, 
\ie, $|\regipca|$  = $|\reginjf|$ = 100, $|\regieval^\textsc{Full}|$  = 229 or $|\regieval^\textsc{Small}|$  = 29.
In each round of Algorithm 1, we sample random 100 among the rest of $\sim 3.5$k (=4000-429) unregistered shapes \unregiset, bring it into correspondence and move it to \regipca when selected by our pruning.
\vspace{-5mm}
\section{Nearest Lookup from Others to GHUM}
\vspace{-1mm}
\begin{figure}[h!]
  \centering
    \includegraphics[width=\linewidth]{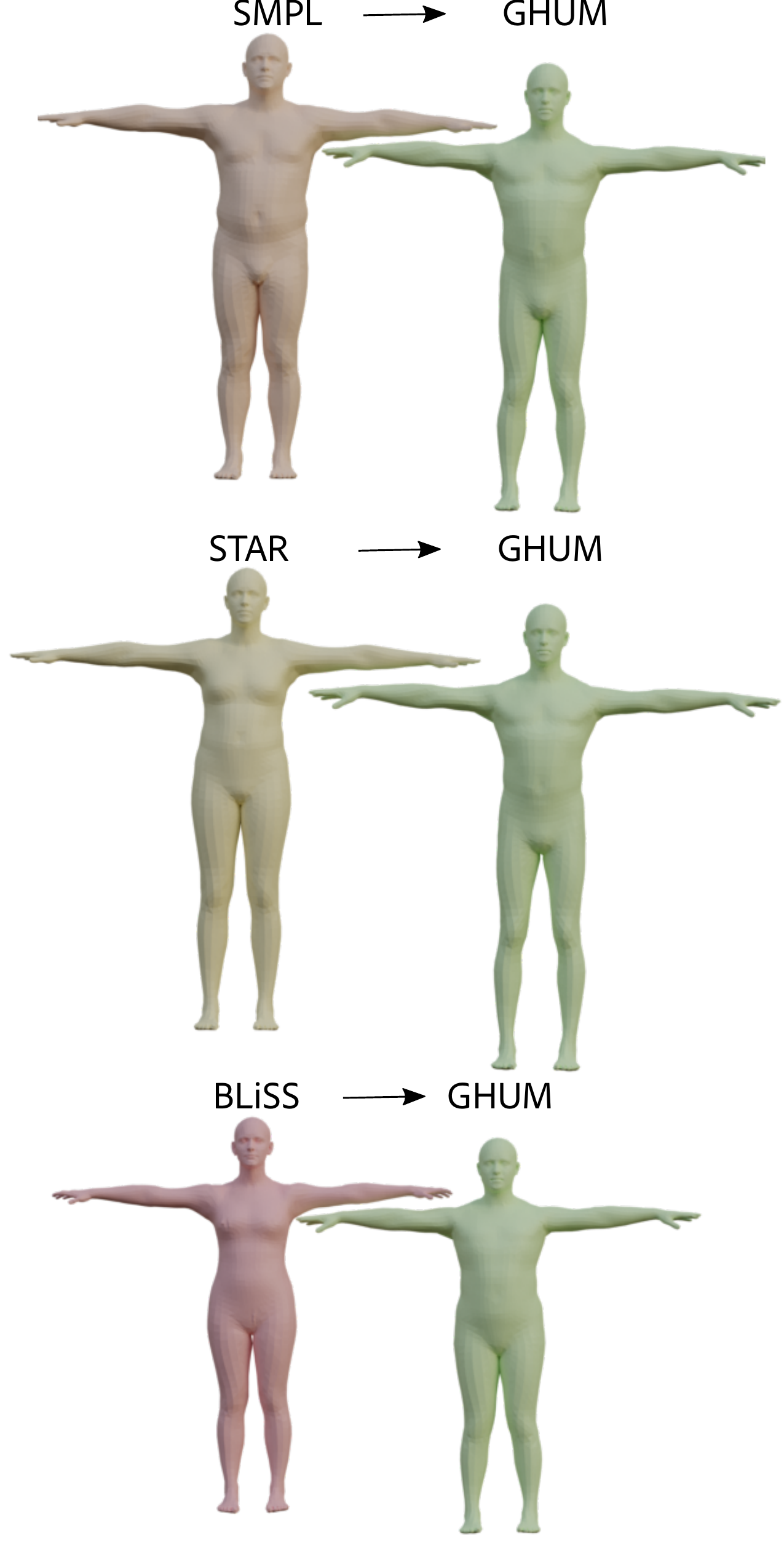}
    \vspace{-2mm}

   \caption{For a sample in each of SMPL, STAR and \name, we lookup the nearest shape in GHUM's space (green). We randomly sampled 10000 shapes in each space.}
   \label{fig:others2ghum}
\end{figure}
In Table 3, we report a v2p error of 2.63 cms on 229 scans, while GHUM \cite{xu2020ghum} report 1.91 cms (Chamfer distance) on the entire CAESAR \cite{CAESAR} dataset. This difference may arise from the difference in the evaluation set and the different training sets. Further, we observe in Table 4 of the main paper, that the numbers reported with respect to GHUM are higher compared to others.  In  Figure \ref{fig:others2ghum}, we show samples from each of SMPL, STAR and \name and their corresponding nearest sample by vertex-to-vertex L2 in GHUM's shape space. Each space was randomly sampled to generate 10000 shapes, and we perform pairwise L2 distance across shapes in different spaces. We observe significant differences in body-shape indicating that there are non-overlapping regions between GHUM and other shape spaces.

\end{document}